\documentclass[10pt,twocolumn,letterpaper]{article}

\usepackage{cvpr}              

\usepackage{amsmath}
\usepackage{amssymb}
\usepackage{booktabs}
\usepackage{enumerate}
\usepackage{mathtools}
\usepackage{bbm}
\usepackage{multirow}
\usepackage{graphicx}

\usepackage[linesnumbered,ruled,vlined]{algorithm2e}
\SetKwInput{KwInput}{Input}                
\SetKwInput{KwOutput}{Output}              

\usepackage[pagebackref,breaklinks,colorlinks]{hyperref}

\usepackage[capitalize]{cleveref}
\crefname{section}{Sec.}{Secs.}
\Crefname{section}{Section}{Sections}
\Crefname{table}{Table}{Tables}
\crefname{table}{Tab.}{Tabs.}

\newcommand{\mC}{\mathcal{C}}
\newcommand{\mL}{\mathcal{L}}

\newcommand{\mF}{\mathcal{F}}
\newcommand{\mI}{\mathcal{I}}
\newcommand{\mS}{\mathcal{S}}
\newcommand{\mD}{\mathcal{D}}
\newcommand{\mG}{\mathcal{G}}
\newcommand{\mP}{\mathcal{P}}

\newcommand{\mE}{\mathcal{E}}
\newcommand{\mV}{\mathcal{V}}
\newcommand{\x}{\boldsymbol{x}}
\newcommand{\e}{\boldsymbol{e}}

\newcommand{\y}{\boldsymbol{y}}
\renewcommand{\Re}{\mathbb{R}}
\renewcommand{\d}{\delta}

\newcommand{\st}{\operatorname{s.t. }}

\newcommand{\npar}[1]{\noindent\textbf{#1}}

\newcommand{\mpar}[1]{\medskip\noindent\textbf{#1}}
\newcommand{\mybullet}{\smallskip\noindent\noindent{-- }}

\def\cvprPaperID{6622} 
\def\confName{CVPR}
\def\confYear{2022}

\begin{document}

\title{Towards Effective Multi-Label Recognition Attacks via Knowledge Graph Consistency}

\author{Hassan Mahmood\\
Northeastern University\\
{\tt\small mahmood.h@northeastern.edu}
\and
Ehsan Elhamifar\\
Northeastern University\\
{\tt\small e.elhamifar@northeastern.edu}
}

\maketitle

\begin{abstract}
Many real-world applications of image recognition require multi-label learning, whose goal is to find all labels in an image. Thus, robustness of such systems to adversarial image perturbations is extremely important. However, despite a large body of recent research on adversarial attacks, the scope of the existing works is mainly limited to the multi-class setting, where each image contains a single label. We show that the naive extensions of multi-class attacks to the multi-label setting lead to violating label relationships, modeled by a knowledge graph, and can be detected using a consistency verification scheme. 
Therefore, we propose a \emph{graph-consistent} multi-label attack framework, which searches for small image perturbations that lead to misclassifying a desired target set while respecting label hierarchies. By extensive experiments on two datasets and using several multi-label recognition models, we show that our method generates extremely successful attacks that, unlike naive multi-label perturbations, 
can produce model predictions consistent with the knowledge graph. 
\end{abstract}
\section{Introduction}
\label{sec:intro}

Despite the tremendous success of Deep Neural Networks (DNNs) for image recognition, DNNs have been shown to be vulnerable to attacks \cite{Goodfellow:ICLR15, Szegedy:ICLR14, Papernot:essp16, Kurakin:ICLR17, Zgner:sigkdd18, Eykholt:arxiv18,  Metzen-Fischer:ICCV19, Li-Tian:ICCV19, Carlini:AISEC17, Hendrycks:CVPR21}. Adversarial attacks are imperceptible image perturbations that result in incorrect prediction with high confidence. They have highlighted the lack of robustness of DNNs and have become a major security and safety concern \cite{Eykholt:CVPR18,Sharif:CCS16, Biggio:kdd13,Thys:cvprw19}. This has motivated a large body of research on generating small imperceptible perturbations 
, and subsequently using the attacks to design robust defense mechanisms, e.g., by detecting attacks or retraining the model using perturbed images. The majority of existing works, however, have focused on multi-class recognition, which assumes that only one class is present in an image \cite{Kurakin:arXiv16, Zheng:arxiv18, yu:cvpr21, FCroce:arxiv20, wong:arxiv20, Athalye:icml18, Dong:cvpr19, gowal:arxiv19}. 

\begin{figure}[t]
    \centering
    \includegraphics[width=8cm]{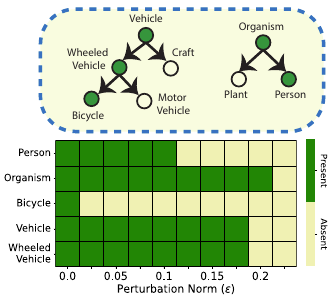}%
     \caption{Generating effective multi-label attacks is challenging. We show the effect of an attack on the label `bicycle' on the output of the multi-label classifier. Top: labels are hierarchically-related. Green nodes show the labels predicted as present by the model before the attack. 
    While the attack misclassifies \emph{bicycle}, 
     it changes the prediction for the \emph{person} for $\epsilon > 0.125$ and also causes inconsistency in the right subgraph (except for the large $\epsilon = 0.225$), which is undesired. For $\epsilon < 0.175$, the \emph{vehicle} and \emph{wheeled vehicle} labels are present despite all descents being absent, causing inconsistency in the left subgraph. 
     }
     \label{fig:fgsmattack}
  \end{figure}

On the other hand, many real-world applications of image recognition require finding all labels in an image. This includes autonomous driving, surveillance and assistive robotics. For example, a self-driving car needs to detect and recognize all objects in the scene or an image may contain multiple human-object interactions. Therefore, multi-label recognition (MLR) aims at finding all labels in an image \cite{Baruch:arxiv20, Yang:cvpr16, Chen:icme19, Ye:ECCV20, Chen:CVPR19, Feng:AAAI19, Jiahao:itm20}. However, MLR is a harder problem than multi-class recognition, as there are combinatorial combinations of labels that could appear in images, while many labels often occupy only small image regions. 

\begin{figure}[t]
    \centering
     \includegraphics[width=6cm]{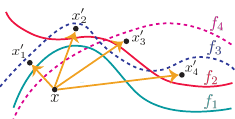}%
     \caption{
     Attacking hierarchically-related labels. Assume that label 2 is the parent of 1 ($f_2 \rightarrow f_1$), label 4 is the parent of 3 ($f_4 \rightarrow f_3$) and the image $\x$ is classified as present by the four classifiers. The goal is to adversarially attack 
     $f_1$ to misclassify $\x$ as being absent.  $\x'_1$ is the 
     adversarial example that is successfully attacked as being absent by $f_1$ but is still classified as present 
     by $f_2$, causing inconsistency in ($f_2 \rightarrow f_1$) hierarchy. $\x'_2$ and $\x'_3$ successfully attacked 
     label 1 and label 2 but cause inconsistency in ($f_4 \rightarrow f_3$) hierarchy. 
     $\x_4'$ shows an attack that misclassifies labels $1$ and $2$ as being absent, while maintaining all label  
     consistencies. 
     }
     \label{fig:efficientattack}
  \end{figure}

Generating adversarially attacks for a multi-label classifier, however, is a challenging problem and different from the multi-class setting. In the multi-class recognition, once a present (or an absent) label is attacked, another label must turn on (or all other labels must turn off). In contrast, in the multi-label setting, we may want to attack and modify several labels at the same time, turning some labels on and some labels off while keeping the remaining labels intact. However, attacks on a set of target labels could drastically modify other labels 
\cite{Yang:ecml21,Yang:aaai21,Song:ICDM18}. 

As we show in Figure \ref{fig:fgsmattack}, for an image that contains `bicycle' and `person', generating an attack to turn off the label `bicycle' results in the undesired attack on `person' for $\epsilon \geq 0.125$. This is mainly because of high correlation among labels that modifying one class may affect unrelated classes. Although the attack successfully turns off the `bicycle', it fails to use the relationships among (bicycle, vehicle, wheeled vehicle) and continues predicting `vehicle' and `wheeled vehicle', rendering the model predictions label inconsistent. We argue that an attack on a set of labels, which leads to changing some or possibly many other labels, can be detected by using a knowledge graph that encodes, e.g., parent-child relationships among labels. For example, `bicycle' and `motor vehicle' are both `wheeled vehicles', therefore, if the label `bicycle' appears in an image with 
`wheeled vehicle' being absent could indicate an adversarial attack, as shown in Figure \ref{fig:efficientattack}. 
In fact, the few existing works on multi-label attacks \cite{Song:ICDM18,Yang:ecml21,Yang:aaai21,Zhou:ijcnn21}, which are simple extensions of multi-class attacks, suffer from violating such label consistencies by not leveraging relationship among labels.

\mpar{Paper Contributions.} We propose a framework for generating effective graph-consistent multi-label attacks, which respect the relationships among labels in a knowledge graph, hence, can stay undetected. We show that the existing naive multi-label attacks lead to violating label relationships and can be detected using an inconsistency verification scheme. Therefore, we propose a formulation that searches for small image perturbations that lead to misclassifying a desired target set while respecting label hierarchies in a knowledge graph. To do so, we search for and use a minimal subset of labels that should be modified in addition to the target set to preserve consistencies according to the knowledge graph. By extensive experiments on two datasets and on several multi-label recognition models, we show that our method generates successful attacks that, unlike naive multi-label perturbations, evade detection.

\begin{figure}[t]
    \centering
     \includegraphics[width=8cm]{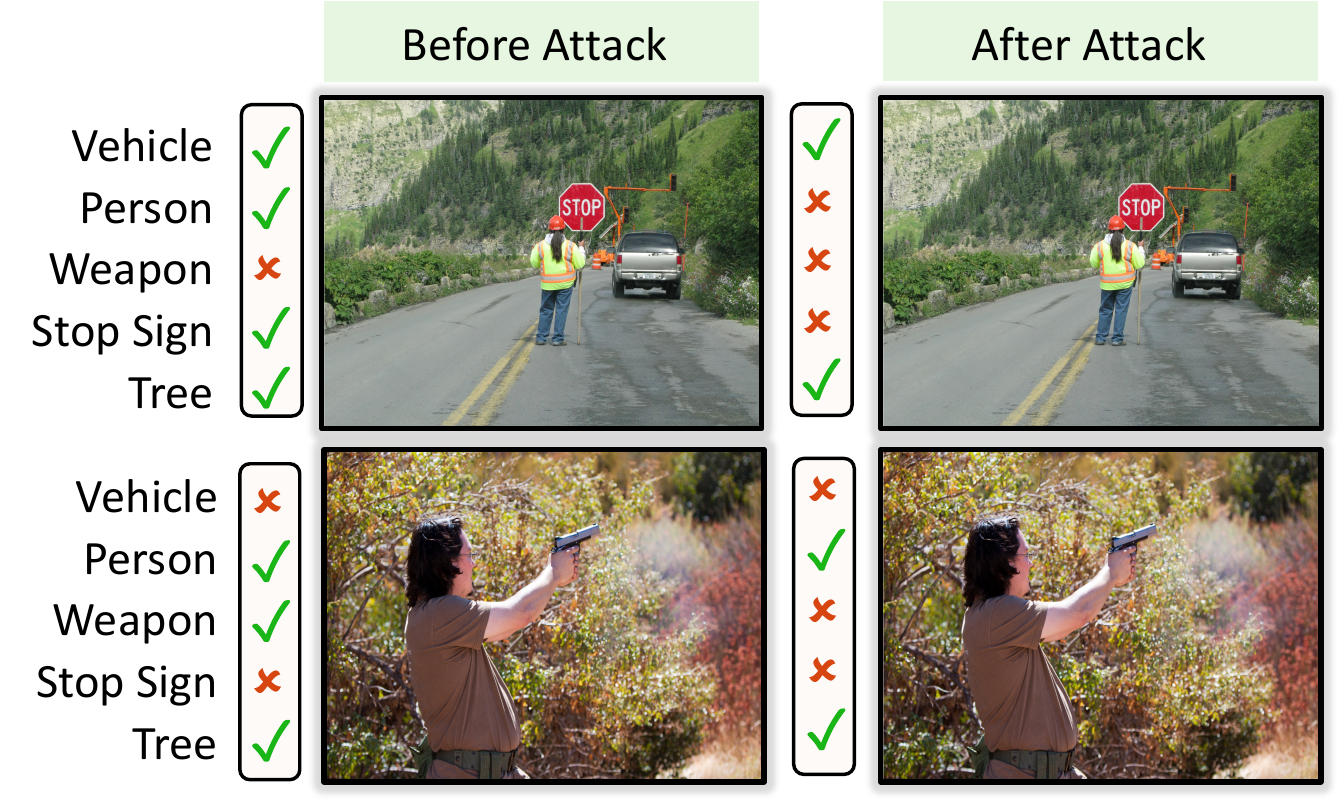}%
     \caption{
     Motivation for the multi-label adversarial attacks. For each image, 
     the predictions of a given multi-label classifier are shown on the left. A label 
     predicted as \emph{present}
     is shown with a tick mark and an \emph{absent} label is shown with a cross mark. 
     The top left image contains trees, a car and a person holding the stop sign on the road. 
     An adversarially perturbed version of this image is shown on the top 
     right. In this case, the classifier does not 
     recognize the person and the stop sign. 
     Similarly, the bottom-left image contains trees and a person holding a gun. An adversarial 
     example for this image (bottom-right) can fool the model to misclassify the \emph{gun} in the image. 
     The figures are taken from OpenImages \cite{OpenImages:16}.  
     }
     \label{fig:mlrexample}
  \end{figure}

\section{Graph-Consistent Multi-Label Recognition Attacks}

\subsection{Problem Setting}

We study generating effective adversarial attacks for the Multi-Label Recognition (MLR) problem. In MLR, multiple labels can appear in an image, see Figure \ref{fig:mlrexample}, as opposed to the multi-class recognition, where each image has only one label. Let $\mC$ denote the set of all labels. For an image $\x \in \Re^d$, the set of its labels can be represented by $\y \in \{-1, +1\}^{|\mC|}$, indicating the presence (+1) or absence (-1) of each label of $\mC$ in the image.  Let $\mF: \Re^d \rightarrow \Re^{|\mC|}$ denote a multi-label classifier, which we assume has already been learned using training images. The multi-label classifier $\mF = \{f_1, f_2, \dots, f_{|\mC|}\}$ consists of $| \mC |$ binary classifiers for each label, where $f_{c}(\x) \in (-\infty,+\infty)$ is the score of the classifier $c$. Therefore, whether the class $c$ is present or absent in the image $\x$ is given by $y_c = \d( f_{c}(\x) )$, where $\d(\cdot)$ is an indicator function, which is +1 when its argument is positive and is -1 otherwise. We denote the set of present labels in the image $\x$ by $\mP(\x)$ and absent labels by $\bar{\mP}(\x)$.

Let $\Omega(\x) \subseteq \mC$ denote the set of labels in the image $\x$ which we want to attack, i.e., after the attacks the present labels in $\Omega(\x)$ must become absent and vice versa. Given the multi-label nature of the problem, crafting attacks can be done in two different ways, which we discuss in Section \ref{sec:mla}. We show the limitation of these attacks and in Section \ref{sec:mla-detection}, discuss an effective detection mechanism based on label hierarchy. In Section \ref{sec:gmla}, we study two effective ways of generating multi-label attacks that leverage relationships among labels and can stay undetected.   

\subsection{Multi-Label Attack (MLA)}
\label{sec:mla}

To generate a multi-label attack on $\x$ that modifies the labels in $\Omega(\x)$, one can take two approaches. First, we can generate a small perturbation $\e \in \Re^d$ by minimizing the \emph{negative} multi-label recognition loss for labels in $\Omega(\x)$ while restricting the magnitude of $\e$. More precisely, we can solve
\begin{equation}
\label{eq:mla-alpha}
\text{MLA$_\alpha$:} ~~ \min_{\e} -\mL_{ce} (\x+\e, \Omega(\x)) ~~ \st ~~ \| \e \|_p \leq \epsilon,
\end{equation}
where $\| \cdot \|_p$ denotes the $\ell_p$-norm and $\mL_{ce} (\x+\e, \Omega(\x))$ is the binary cross entropy loss on target labels for the perturbed image and is defined as
\begin{equation}
\label{eq:bce-alpha}
\small
\sum_{c \in \Omega(\x)} \!\! y_c \log \sigma (f_c(\x+\e)) + (1-y_c) \log(1-\sigma(f_c(\x+\e))).
\end{equation}
This type of attack, which we refer to as $\text{MLA}_\alpha$ tries to find a perturbation $\e$ that flips the labels in $\Omega(\x)$. However, the optimal perturbation could simultaneously change many of the other labels in $\mC \backslash \Omega(\x)$, which is undesired.

A second type of attack, which we refer to as $\text{MLA}_\beta$, therefore, finds a small perturbation $\e$ that changes the labels in $\Omega(\x)$ while keeping all other labels, i.e., the labels in $\mC \backslash \Omega(\x)$, fixed. One can achieve this by solving 
\begin{equation}
\label{eq:mla-beta}
\begin{split}
\text{MLA$_\beta$:} ~~ &\min_{\e} -\mL_{ce} (\x+\e, \Omega(\x)) + \mL_{ce} (\x+\e, \mC \backslash \Omega(\x)),\\ &~\st ~~ \| \e \|_p \leq \epsilon,
\end{split}
\end{equation}
where the first term in the objective function tries to flip the labels in $\Omega(\x)$ while the second term preserves the labels in $\mC \backslash \Omega(\x)$. Notice that with the additional objective, the space of perturbations for $\text{MLA}_\beta$ is smaller than that of $\text{MLA}_\alpha$, yet it ensures not modifying labels outside the target set. 

As we discuss next, we can detect both types of above attacks by using relationships among labels, which would make them almost ineffective. This motivates us to propose a new class of MLR attacks that leverages the relationships among labels to not only change the labels in the target set, but can stay undetected.

\subsection{MLA Detection using Knowledge Graphs}
\label{sec:mla-detection}

Let $\mG = (\mC, \mE)$ denote a directed acyclic knowledge graph built on the labels $\mC$, where $\mE$ denotes the set of edges (see below for the details about building this graph). We propose to use the parent-child label relationship in the directed acyclic graph to search for and detect inconsistencies in label prediction after an attack. In particular, given an image, the output of the MLR is consistent if and only if \emph{i)} when the classifier predicts +1 for a parent node/label, then at least one of its children must also be predicted as +1; \emph{ii)} when all children of a node/label are predicted as -1, then the parent must be predicted as -1. We study two ways to check for such label consistency via $\mG$.
\begin{enumerate}[1)]
    \item \emph{Local Consistency Verification}: For a target label $c$, we consider a local subgraph of $\mG$ around $c$ and check for predicted label consistency in the subgraph, to determine whether an attack has occurred near $c$. 
    \item \emph{Global Consistency Verification}: We check if the entire graph $\mG$ is label consistent for the output of the multi-label classifier. Satisfying global consistency is harder. 
\end{enumerate}

As we show in the experiments, both detection methods can successfully detect attacks for the $\text{MLA}_\alpha$ and $\text{MLA}_\beta$ perturbations, which we discussed in Section \ref{sec:mla}. In the next subsection, we leverage the label-relationships from the knowledge graph to craft successful yet undetectable attacks. 

\mpar{Building the Knowledge Graph.} To construct the hierarchical knowledge graph, we use WordNet \cite{Miller:ACM95}, which contains rich semantic relationships between labels\footnote{WordNet is a lexical database for the English language, containing 155,327 words organized in 175,979 synsets.} (one could also use other sources, such as ConceptNet \cite{Speer:AAAI17}). We build a tree $\mG = (\mC, \mE)$ on the labels $\mC$, where $\mE$ denotes the set of edges. 

For each label in $\mC$, we use hypernym and hyponym relations in the WordNet to extract its parent and child labels. For example, for the label `car', we obtain `vehicle' as the parent using its hypernyms. Since a word can be associated with several synsets in the WordNet, we choose the synset with the closest match to the description of the label. To build the tree, we use the maximum WUP similarity score \cite{Pedersen:naacl04} between a child node and multiple parents to select a single parent node.


\begin{figure}[!tbp]
    \centering
  \begin{subfigure}[b]{0.11\textwidth}
    \includegraphics[width=\textwidth]{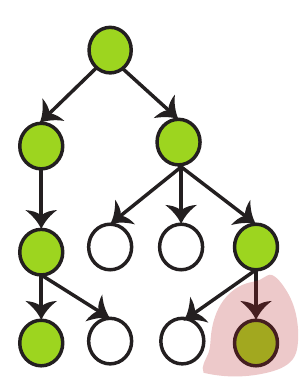}
    \caption{MLA$_\alpha$}
    \label{fig:mlaalpha}
  \end{subfigure}
  \begin{subfigure}[b]{0.11\textwidth}
    \includegraphics[width=\textwidth]{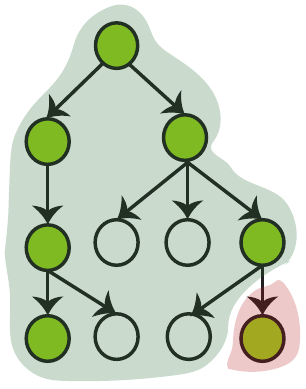}
    \caption{MLA$_\beta$}
    \label{fig:mlabeta}
  \end{subfigure}
  \begin{subfigure}[b]{0.11\textwidth}
    \includegraphics[width=\textwidth]{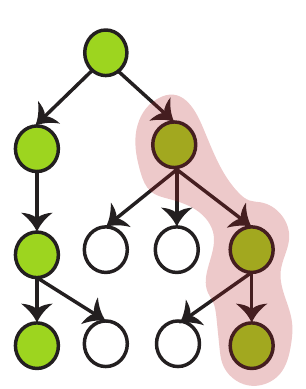}
    \caption{GMLA$_\alpha$}
    \label{fig:gmlaalpha}
  \end{subfigure}
  \begin{subfigure}[b]{0.11\textwidth}
    \includegraphics[width=\textwidth]{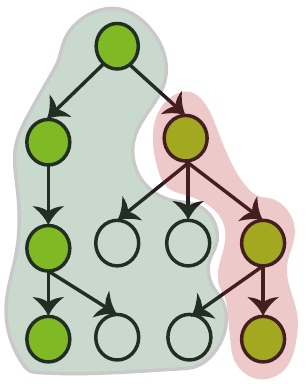}
    \caption{GMLA$_\beta$}
    \label{fig:gmlabeta}
  \end{subfigure}
  \label{fig:attacks}
  \caption{Four different ways of generating multi-label attacks. The green nodes show the \emph{present} labels predicted by the model. The labels to be modified $\Gamma$ are shown within the red region and the 
  labels to be fixed $\mC \backslash \Gamma$ are shown within the green region.}
    \label{fig:attacks-mlltypes}
\end{figure}

%
\subsection{Graph-based Multi-Label Attack (GMLA)}
\label{sec:gmla}

To find a multi-label attack that can evade detection using the knowledge graph, we propose to find a minimal set of labels whose predictions must be modified to ensure label consistency according to the graph. More specifically, our goal is to design a function $h(\cdot)$ whose input is the target set $\Omega(\x)$ and the knowledge graph $\mG$ and whose output is the expanded target set $\Gamma(\x)$ that needs to be modified for graph consistency. We use the word `ON' for a node predicted as +1 and `OFF' for a node predicted as -1. 

Algorithm \ref{alg:gmlaalgo} shows the steps of our algorithm to obtain such a desired set. 
The algorithm works roughly as follows. Given a set of target labels $\Omega$ present in an image, predicted labels $\mS$, and  the adjacency matrix $\mE$ of knowledge graph, the algorithm finds the minimum set of related labels to be flipped for all target labels in $\Omega$, such that the local consistency of the graph is maintained. While attacking a label, we need to maintain its consistency with respect to its children and parents. To maintain children consistency, each child of the target node must be turned OFF unless that child has multiple parents ON. We parse the path from target node to the leaf nodes and perform the same operation on every node. Similarly, to maintain parents consistency, all parents must be turned OFF unless some parent has more than one child ON. We perform this process for each node along the path from target node to the root node until we do not find more nodes to be attacked.

Using the Algorithm \ref{alg:gmlaalgo} for finding $\Gamma(\x)$, we consider two types of graph-based multi-label attacks (GMLAs). First,
\begin{equation}
\label{eq:gmla-alpha}
\begin{split}
\text{GMLA$_\alpha$:} ~~ &\min_{\e} -\mL_{ce} (\x+\e, \Gamma(\x)) \\
&~\st ~~ \| \e \|_p \leq \epsilon,~~\Gamma(\x) = h\big(\Omega(\x), \mG\big),
\end{split}
\end{equation}
which tries find a perturbation $\e$ that changes the labels in the expanded target set $\Gamma(\x)$. However, the optimal perturbation could potentially change other labels in $\mC \backslash \Gamma(\x)$, which may not be desired. Therefore, we also study
\begin{equation}
\label{eq:gmla-beta}
\begin{split}
\text{GMLA$_\beta$:} ~~ &\min_{\e} -\mL_{ce} (\x+\e, \Gamma(\x)) + \mL_{ce} (\x+\e, \mC \backslash \Gamma(\x)),\\ &~\st ~~ \| \e \|_p \leq \epsilon,~~\Gamma(\x) = h\big(\Omega(\x), \mG\big),
\end{split}
\end{equation}
which finds a small perturbation $\e$ that changes the labels in $\Gamma(\x)$ while keeping all other labels, i.e., the labels in $\mC \backslash \Gamma(\x)$. As we show in the experiment $\text{GMLA}_\beta$ is the most effective type of multi-label attack: it ensures that the prediction of the model across all labels change in a way that the entire predictions would be consistent w.r.t. the graph. Figure \ref{fig:attacks-mlltypes} shows a comparison of different multi-label attacks. To compute these four types of attacks, we \emph{use iterative Projected Gradient Descent (PGD)} \cite{Madry:ICLR18, Kurakin:arXiv16} that iteratively uses first-order gradient of the objective function to compute the adversarial example within an $\epsilon$-norm ball.

\begin{algorithm}[h]
\small
\DontPrintSemicolon
\KwInput{$\Omega$: Target Set, $\mS$: Label Predictions, \; \hskip3em $\mE$: Knowledge Graph's Adjacency Matrix}
\KwOutput{$\Gamma$: Expanded Target Set}
\textbf{Procedure:} $f_{select}(X)$: $\KwRet \hskip0.3em \{i : X_i = True\}$
\;
\newcommand{\var}{\texttt}
\SetKwFunction{FPN}{Process\_Node}
\SetKwFunction{Fchildren}{$f_{child.}$}
\SetKwFunction{Fparents}{$f_{par.}$}
\SetKwProg{Fn}{Procedure}{:}{}
\Fn{\Fchildren{$n, \mE, \mS$}}{
\KwRet $f_{select}(\mE_{[n,:]} \odot \mS == 1$)
}

\SetKwProg{Fn}{Procedure}{:}{}
\Fn{\Fparents{$n, \mE, \mS$}}{
\KwRet $f_{select}(\mE_{[:,n]} \odot \mS == 1$) 
}

\SetKwProg{Fn}{Procedure}{:}{}
\Fn{\FPN{$n, V, \mE, \mS, \Gamma, f_1, f_2$}}{
    Queue $\mathcal{Q}$;\;
    $I \gets f_1(n, \mE, \mS)$ \;
    $\mathcal{Q}.enqueue(I$) \;
    \While{$\mathcal{Q}$ is not empty}{
        $v_n = \mathcal{Q}.dequeue()$\;
        \If{$v_n \notin \mV $}
        {
            $\mV \gets \mV \cup \{v_n\}$ \;
            $I \gets f_2(v_n, \mE, \mS) \backslash \Gamma $\;
            \If{$|I|\;<\;2$}
            {
                $\Gamma \gets \Gamma \cup \{v_n\}$ \;
                $I \gets f_1(v_n, \mE, \mS)$\;
                $\mathcal{Q}.enqueue(I)$\;    
            }}}}
$\Gamma = \{\} $\;

\ForEach{$n \in \Omega$}{
    $\mV = \{n\}$ \;
    $\Gamma \leftarrow$ Process\_Node($n, \mV, \mE, \mS, \Gamma, f_{child.}, f_{par.}$)\;
    $\Gamma \leftarrow$ Process\_Node($n, \mV, \mE, \mS, \Gamma, f_{par.}, f_{child.}$)\;
}
\caption{Graph-based Multi-Label Attack}
\label{alg:gmlaalgo}
\end{algorithm}

\begin{figure*}
    \centering
    \setlength\tabcolsep{0.4pt}
    \begin{tabular}{cc@{\hskip 0.15cm}@{\hskip 0.15cm}cc}
    \multicolumn{2}{c}{PASCAL-VOC} & \multicolumn{2}{c}{NUS-WIDE} \\
    \includegraphics[width = 1.6in]{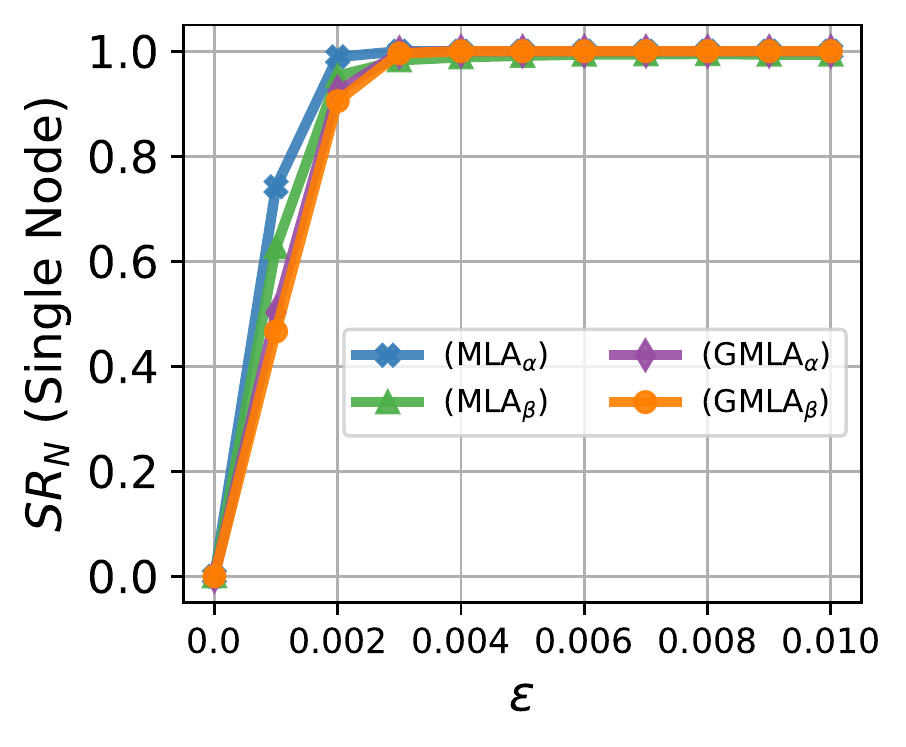} &
    \includegraphics[width = 1.6in]{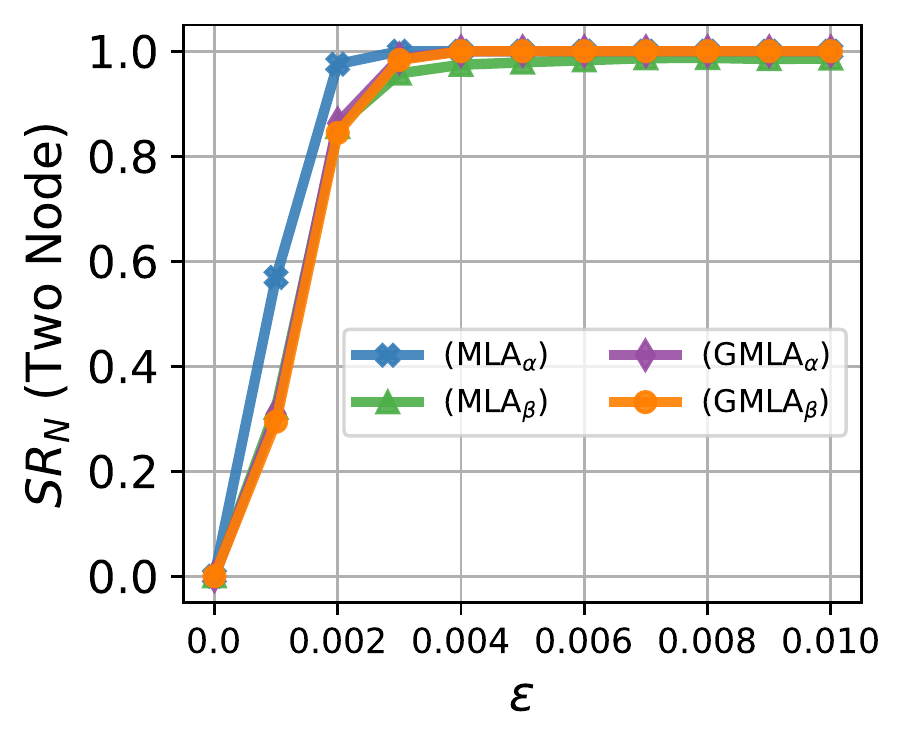} &
    \includegraphics[width = 1.6in]{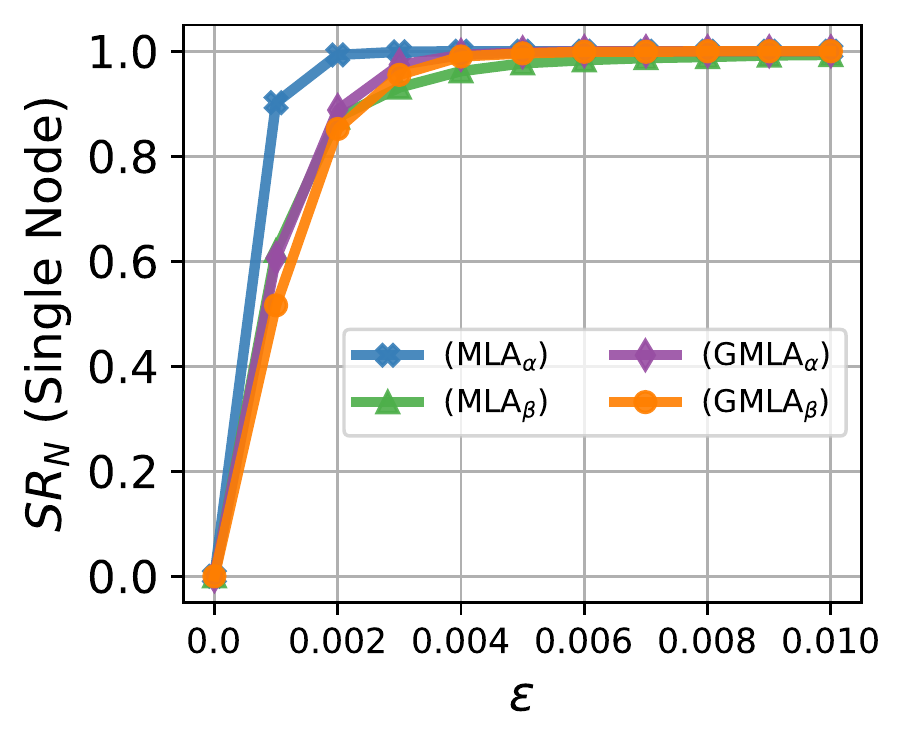} &
    \includegraphics[width = 1.6in]{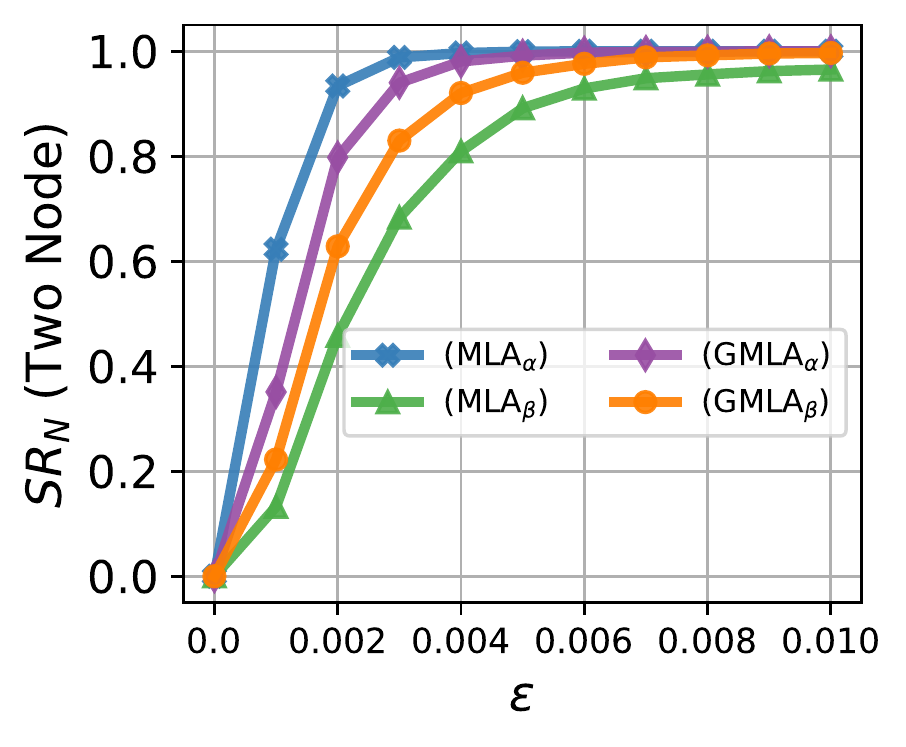} \\

    \includegraphics[width = 1.6in]{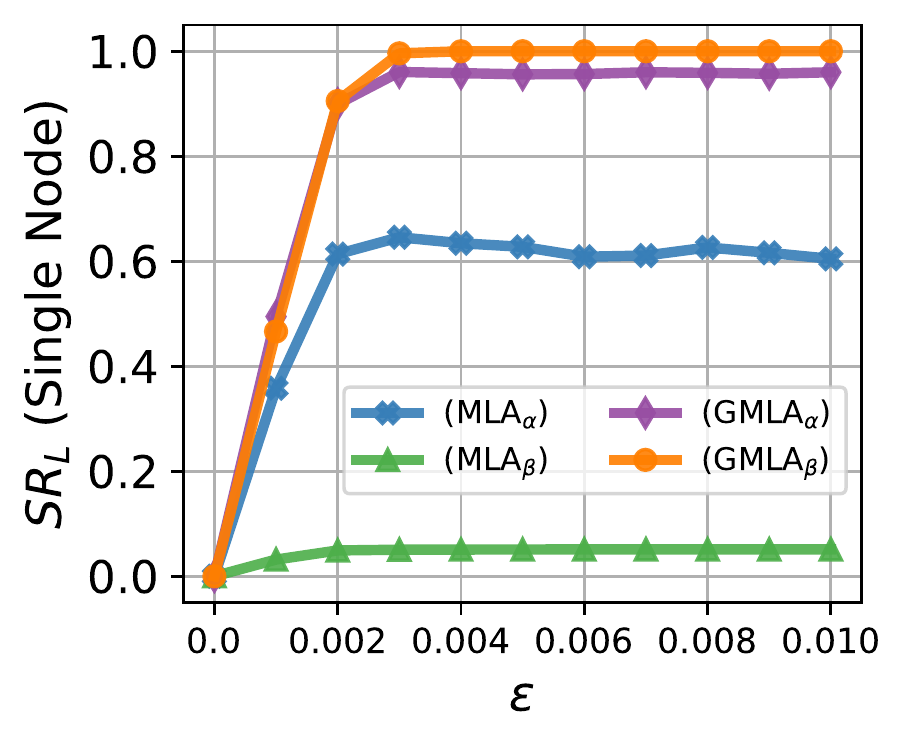} &
    \includegraphics[width = 1.6in]{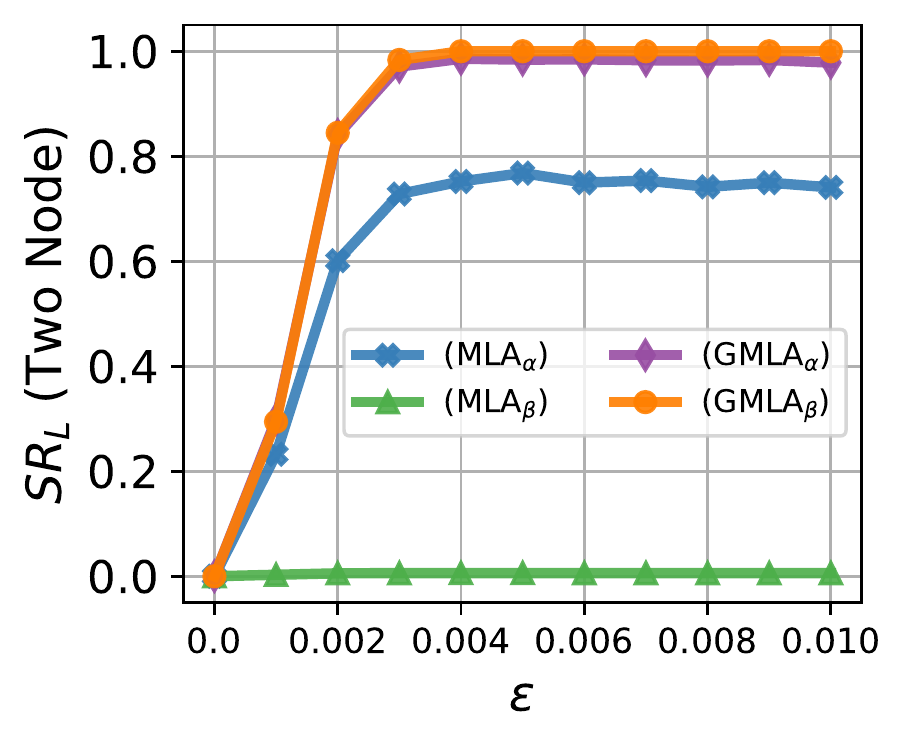} &
    \includegraphics[width = 1.6in]{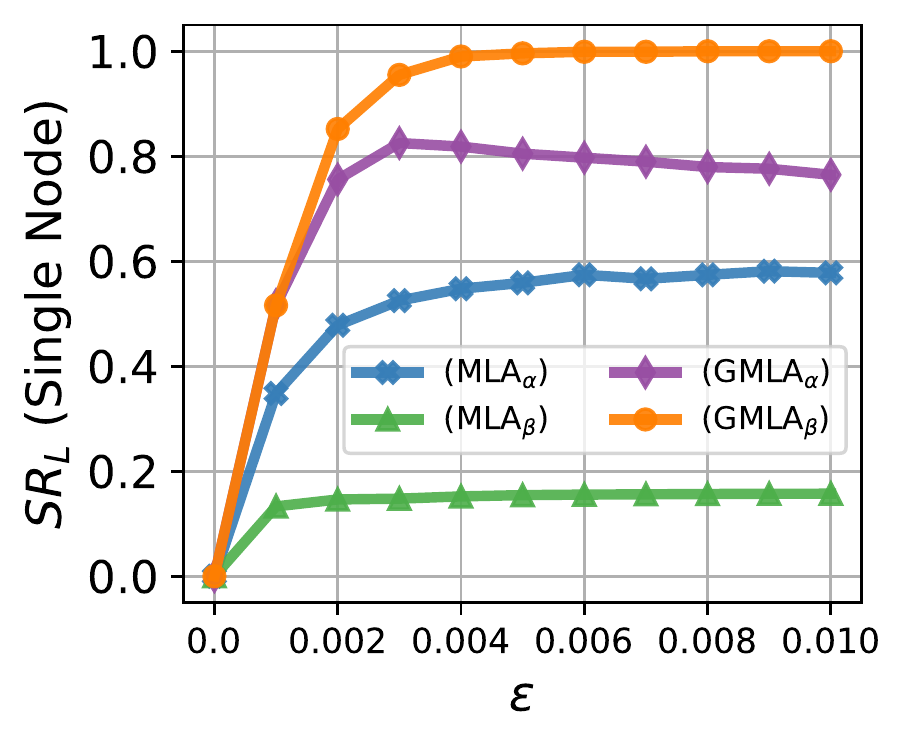} &
    \includegraphics[width = 1.6in]{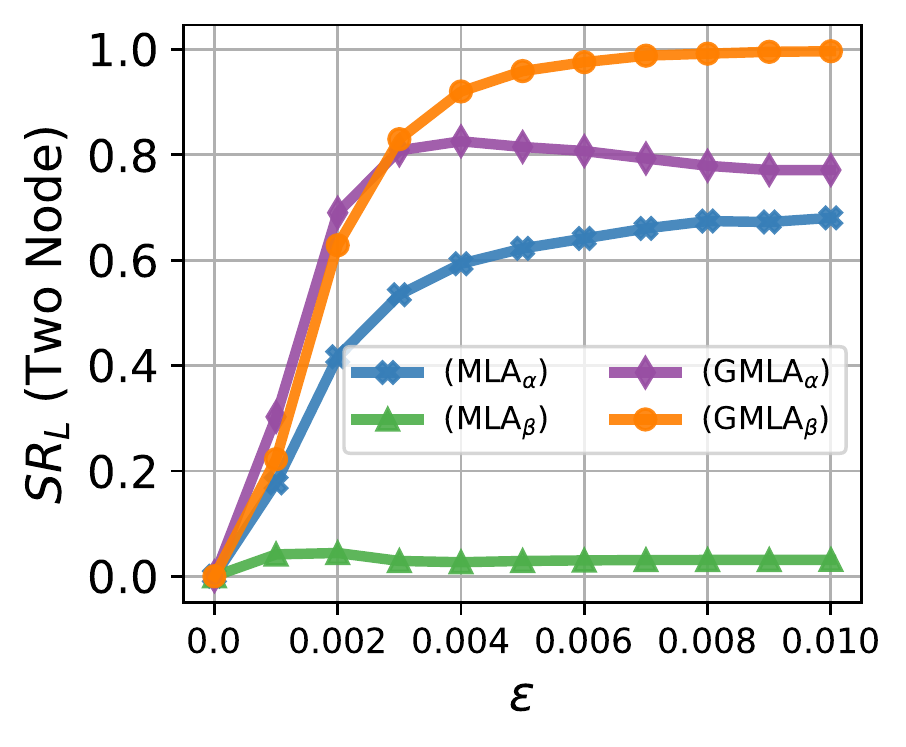} \\

    \includegraphics[width = 1.6in]{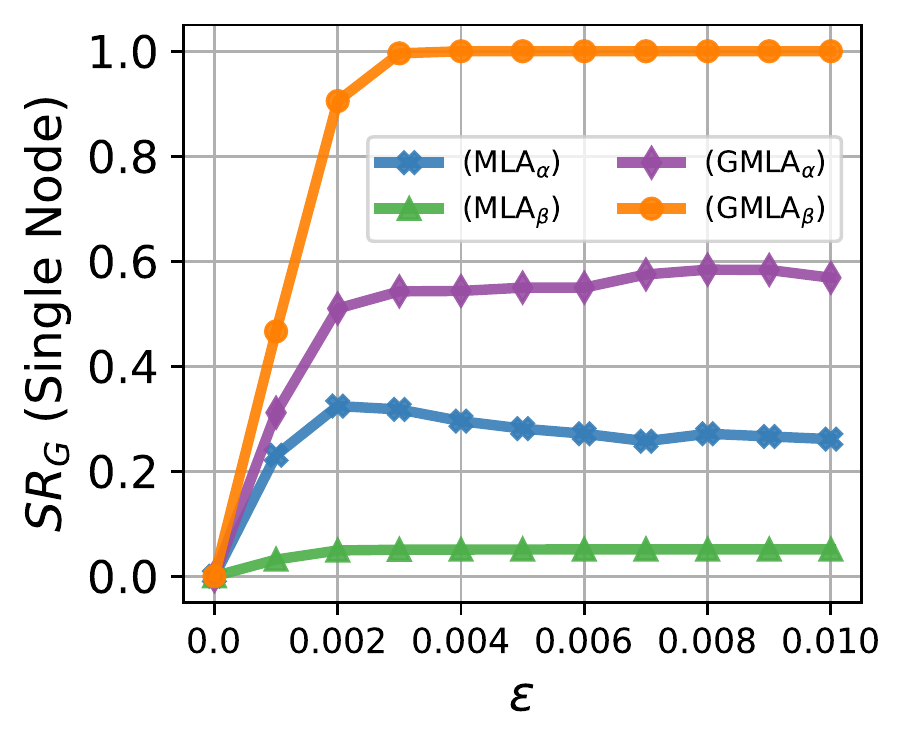}&
    \includegraphics[width = 1.6in]{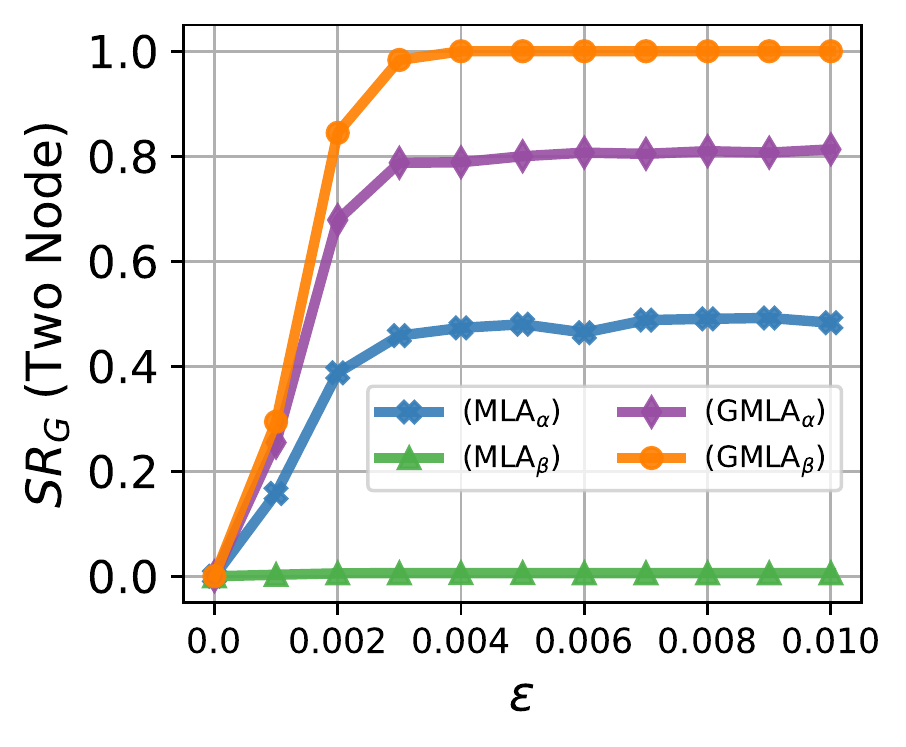}&
    \includegraphics[width = 1.6in]{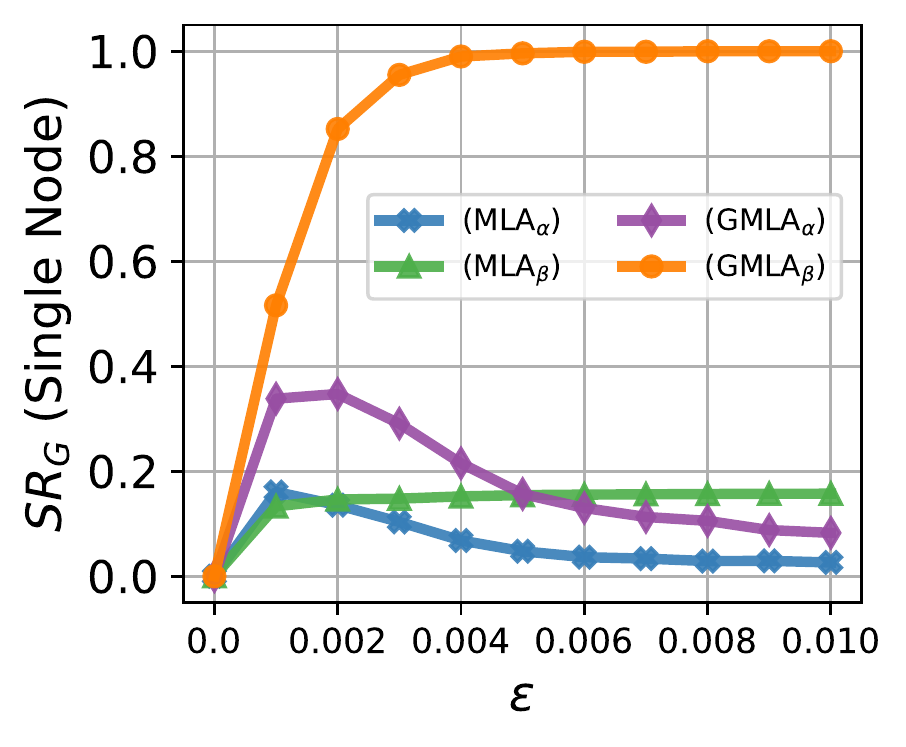}&
    \includegraphics[width = 1.6in]{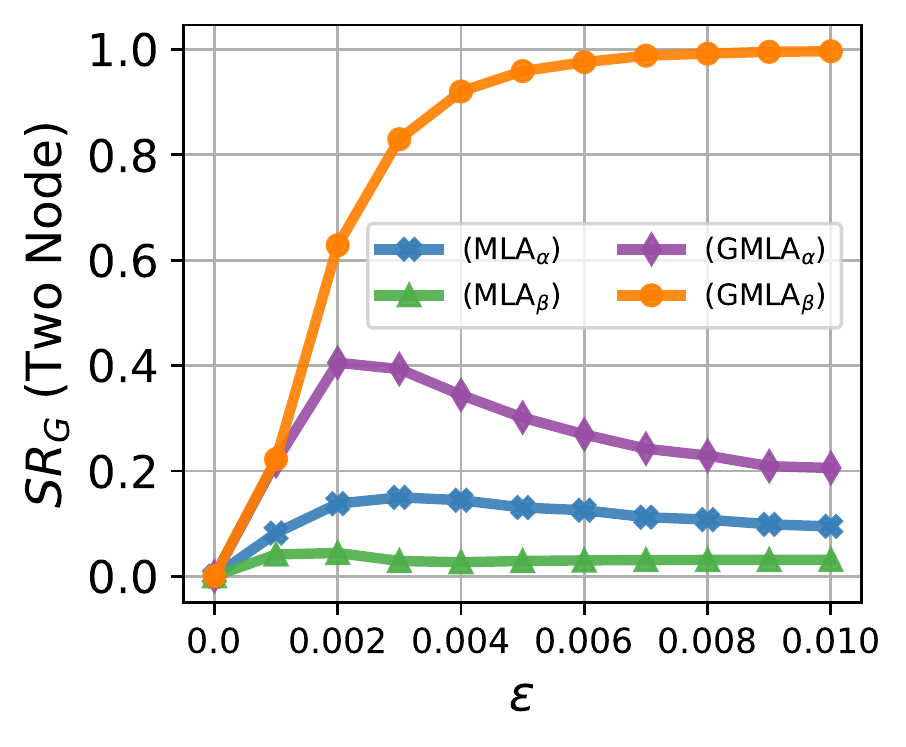}\\
    \end{tabular}
    \vspace{-3mm}
    \caption{\small Success rate of four types of attacks on PASCAL-VOC and NUS-WIDE. 
    The x-axis shows the upper bound on the $l_{\infty}$-norm of perturbations ($\epsilon$). Top: naive attack success rate ($SR_N$). Middle: attack success rate after local verification ($SR_L$). Bottom: attack success rate after global verification ($SR_G$).}
    \label[]{fig:attacksuccess}
\end{figure*}
%

\vspace{-4mm}
\section{Experiments}

\subsection{Experimental Setup}

\npar{Datasets.} We used the Pascal-VOC \cite{pascalvoc} and NUS-WIDE \cite{NUS-Wide:09} for studying the effectiveness of multi-label attacks. 
For Pascal-VOC, we trained each MLR model on 8000 images from the training sets of PASCAL-VOC 2007 
and PASCAL-VOC 2012 and created the adversarial examples for the test set of PASCAL-VOC 2007. We extracted abstract classes from WordNet and built a hierarchical graph among labels. The original 20 labels were expanded into 35 labels after the inclusion of abstract classes. For NUS-WIDE, we trained each MLR model on 150K images from the training set and attacked the models using the test set of the dataset. We used Wordnet to extract abstract classes and built a tree on labels. The total number of labels are 116, which includes 80 original labels and 36 additional abstract classes from WordNet.


%
\begin{figure*}
    \centering
    \setlength\tabcolsep{0.4pt}
    \begin{tabular}{cc@{\hskip 0.15cm}@{\hskip 0.15cm}cc}
    \multicolumn{2}{c}{PASCAL-VOC} & \multicolumn{2}{c}{NUS-WIDE} \\
    \includegraphics[width = 1.6in]{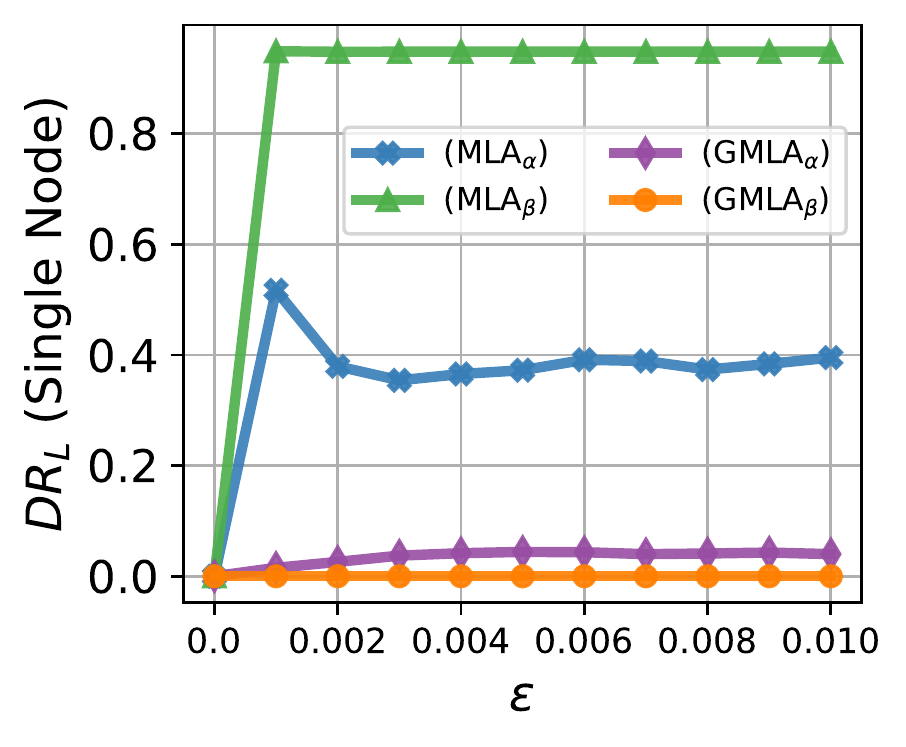} &
    \includegraphics[width = 1.6in]{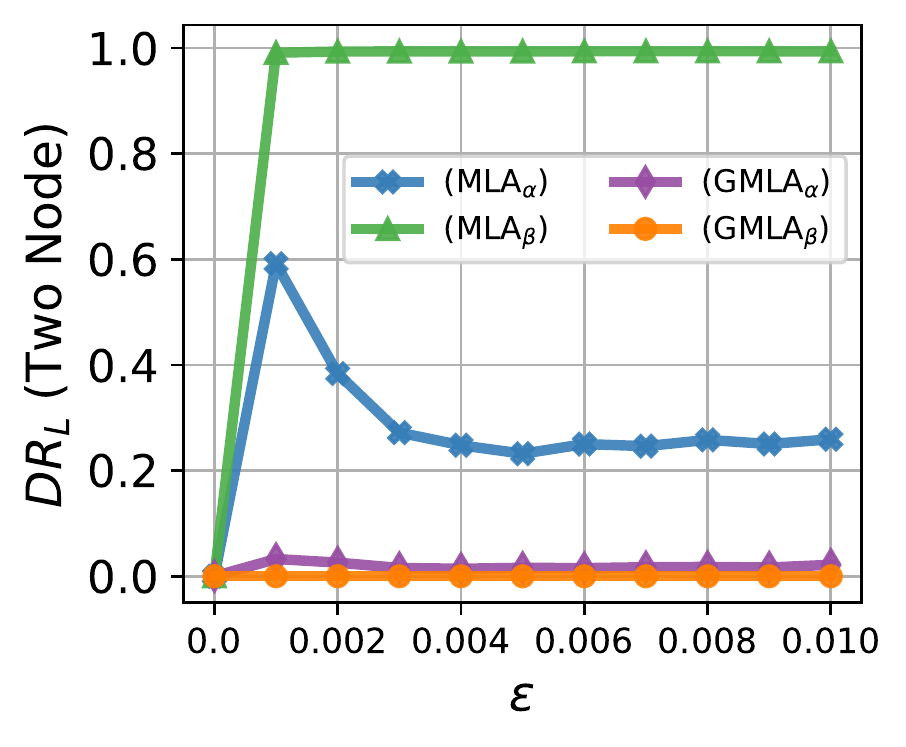} &
    \includegraphics[width = 1.6in]{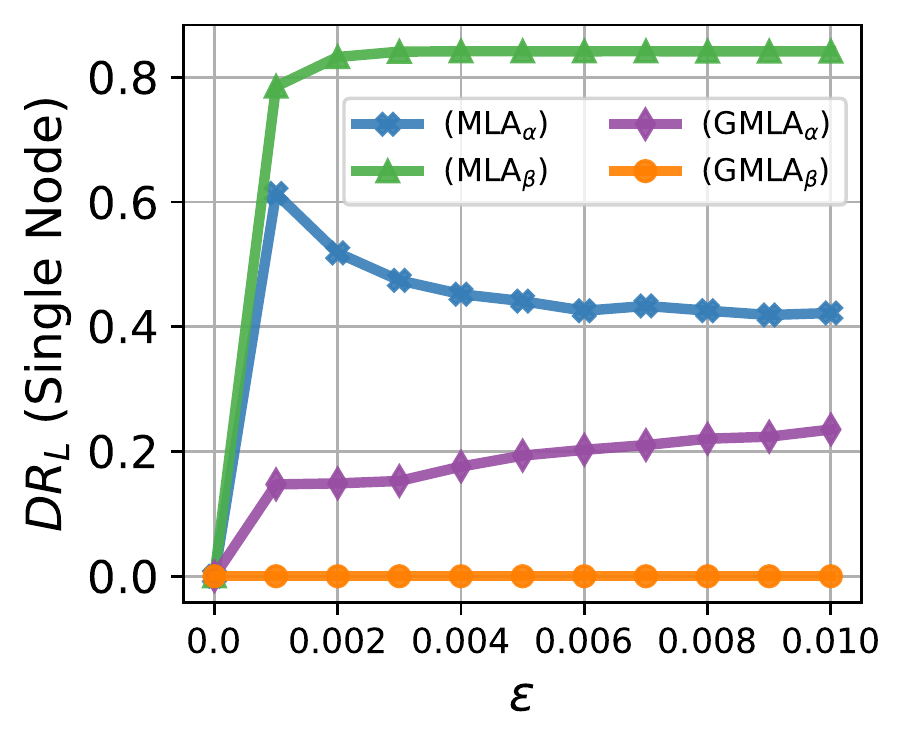} &
    \includegraphics[width = 1.6in]{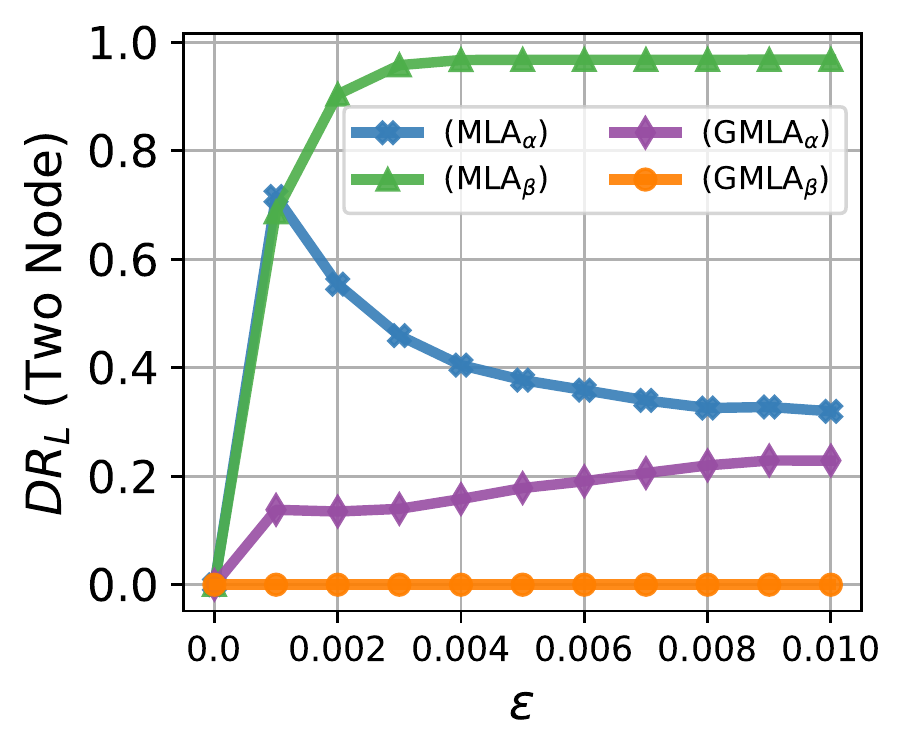} \\

    \includegraphics[width = 1.6in]{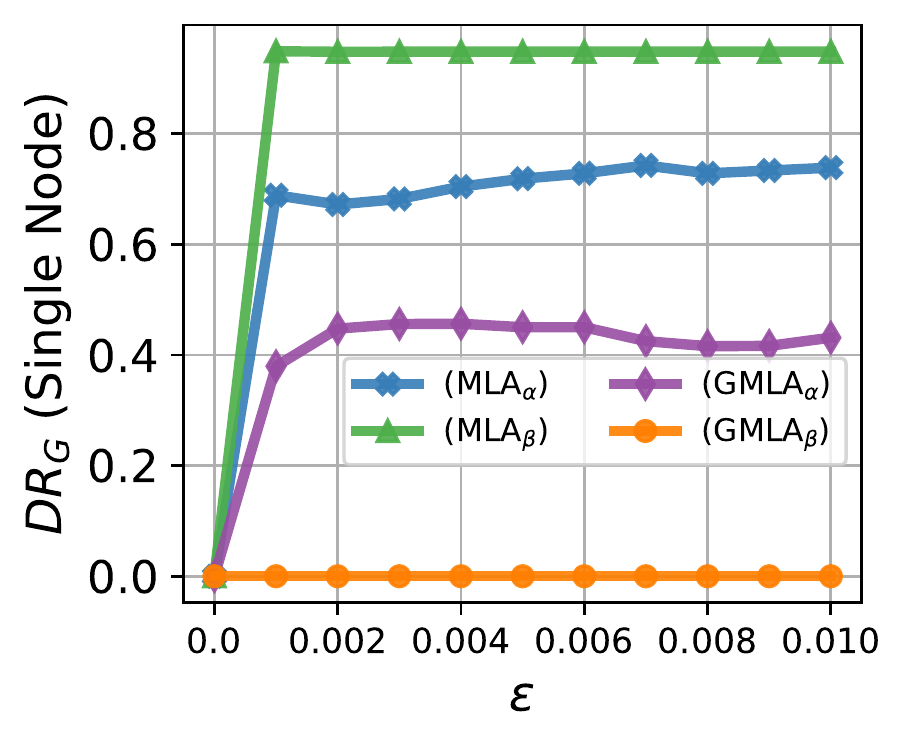} &
    \includegraphics[width = 1.6in]{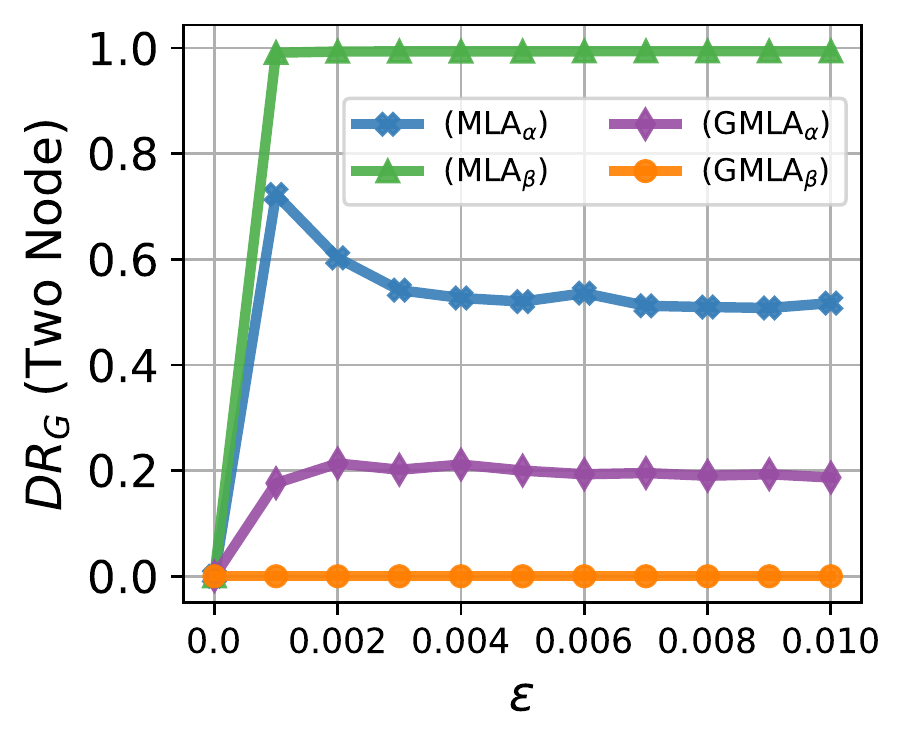} &
    \includegraphics[width = 1.6in]{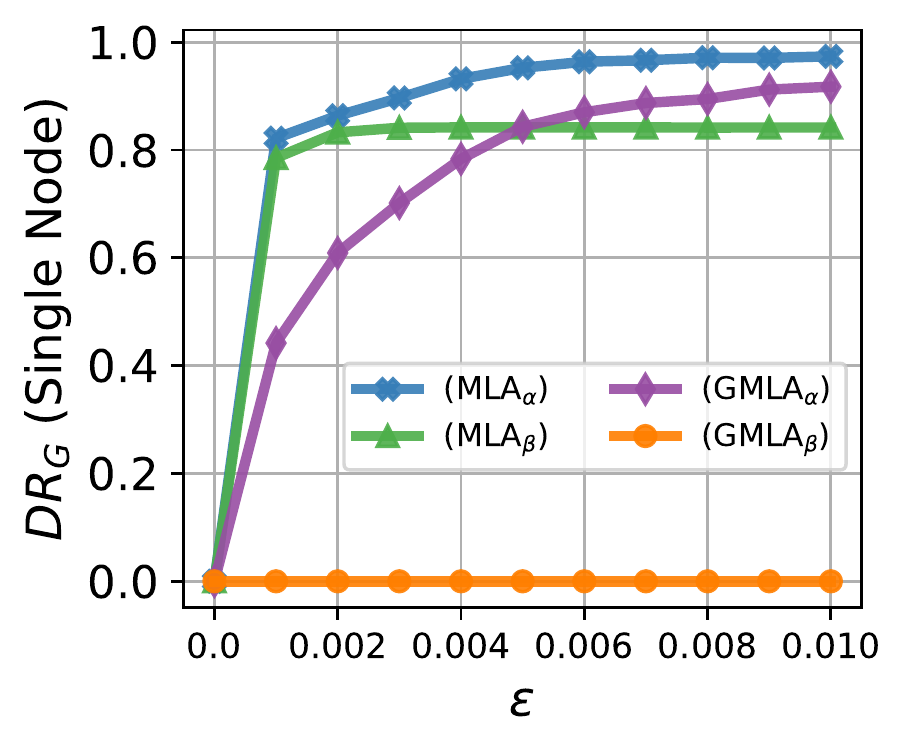} &
    \includegraphics[width = 1.6in]{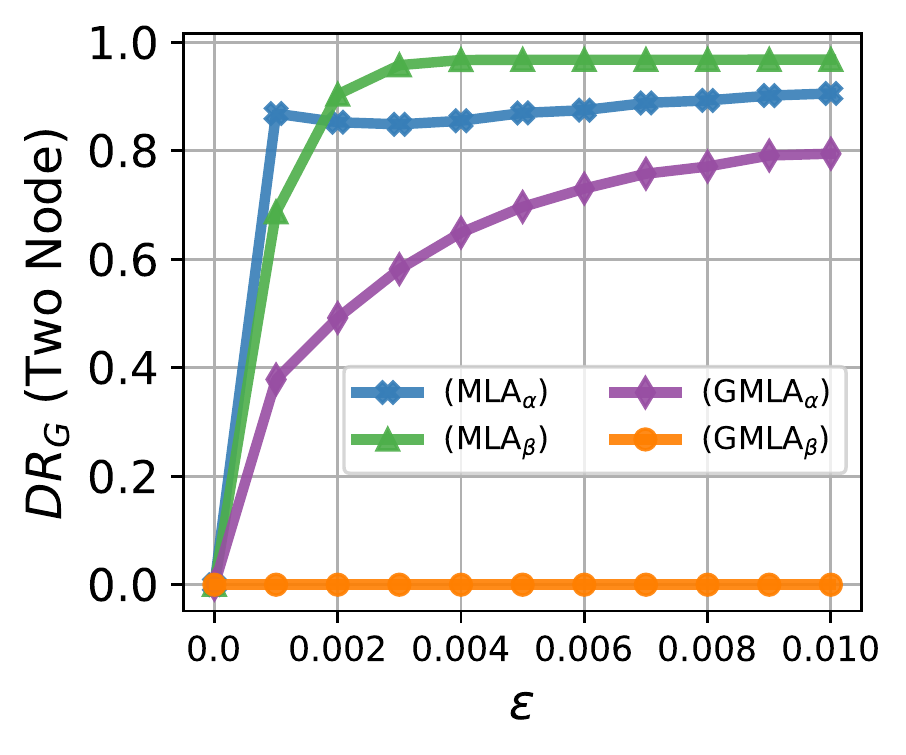} \\
    \end{tabular}
    \vspace{-3mm}
    \caption{\small Detection rate of four types of attacks on PASCAL-VOC and NUS-WIDE. 
    The x-axis shows the upper bound on the $l_{\infty}$-norm of perturbations ($\epsilon$). Top: using local consistency verification ($DR_L$). Bottom: using global consistency verification ($DR_G$).}
%
%
    \label[]{fig:attackdetect}
\end{figure*}

\mpar{Multi-Label Recognition Models.} 
\label{sec:mlrmodels}
We investigated the effectiveness of multi-label attacks on three MLR models, which we trained and attacked on both datasets. 
We use \emph{Binary Relevance} \cite{Tsoumakas:IJDWM07}, which is the simplest extension of the multi-class classification to perform multi-label classification using independent binary classifiers, one for each label. We used the pretrained ResNet101 as the backbone for image-level feature extraction followed by a class-specific feature extraction module, similar to the one used in \cite{Dao:arxiv21}. The class-specific features are then passed to independent binary classifiers. We used the binary cross-entropy loss to train the model. Note that this model does not use any explicit mechanism to learn or leverage label relationships. 

We also use the recent work on \emph{Asymmetric Loss} \cite{Baruch:arxiv20} for multi-label learning that uses a novel asymmetric loss for better optimization over highly imbalanced positive and negative class distributions. Following the experimental setting of \cite{Baruch:arxiv20}, we trained the TResNet-L \cite{Ridnik:arxiv20} backbone. 

Finally, we use \emph{ML-GCN} \cite{Chen:CVPR19} that explicitly learns relationships among labels using Graph Convolutional Networks (GCN). It builds a graph using the word embeddings and the co-occurrence matrix of labels and uses a GCN to extract information about label relationships. We trained the model with the binary cross-entropy loss.

\mpar{Evaluation Metrics.} 
We use two types of evaluation metrics to measure the success of attacks and detection of attacks. Let $\mI$ be the set of images that we attack and $\mS \subseteq \mI$ denote the set of images that are successfully attacked, i.e., for $\x \in \mS$, all labels in $\Omega(\x)$ change after the attack. 

We define the \emph{local and global detection rates}, denoted by $DR_L$ and $DR_G$, respectively, as the the percentage of successfully attacked images ($\mS$) for which we can detect the attack by the local and global consistency verification methods, discussed in Section \ref{sec:mla-detection},
\begin{equation}
DR_L = \frac{|\mD_L|}{|\mS|},~~DR_G = \frac{|\mD_G|}{|\mS|},
\end{equation}
where $\mD_L$ and $\mD_G$ denote the subsets of $\mS$ for which the attack is detected using, respectively, the local and global consistency verification.

We also measure three different success rates for the attacks. First, we compute the naive success rate $SR_N$, which denotes the percentage of all attacked images whose attack was successful. We also compute the local and global success rates, denoted by $SR_L$ and $SR_G$, respectively, which denote the percentage of attacked images whose attack is successful and cannot be detected by the local and global consistency verification, i.e., 
\vspace{-1mm}
\begin{equation}
SR_N = \frac{| \mS |}{|\mI|},~~SR_L = \frac{| \mS \backslash \mD_L|}{|\mI|},~~SR_G = \frac{| \mS \backslash \mD_G|}{|\mI|}.
\vspace{-1mm}
\end{equation}

\mpar{Implementation Details.} We run different attacks by using $p = \infty$, i.e., 
using $\| \e \|_{\infty}$. We study \emph{single-node} attack, where we randomly 
select a single present label in the output of the MLR model to attack, 
hence, $| \Omega(\x) | = 1$ for every $\x$. We also study the \emph{multi-node} 
attack, where we randomly select multiple present labels in the MLR output to attack. 
In the main paper, we show results for one- and two-node attacks. We implemented all models in Pytorch and train them until the validation loss 
converges. We use ADAM optimizer\cite{Kingma:ICLR15} for \emph{Binary-Relevance} and 
\emph{TResNet-L} models; and SGD optimizer with momentum = $0.9$, weight decay = $1\mathrm{e}^{-4}$ for ML-GCN. 
We use learning rate of $10^{-4}$ for PASCAL-VOC and NUS-WIDE with a batch size of 150 and 70, respectively. 
For all models, we reduce the learning rate by $0.1$ at epochs 25, 100, 300 for PASCAL-VOC and at epochs 25, 50, 100, 200 for NUS-WIDE.

\subsection{Experimental Results}
In this section, we investigate the success rates of different multi-label attacks as well as the performance of the two attack detection methods using the three multi-label recognition models, discussed in the previous part.

Figure \ref{fig:attacksuccess} shows success rates of different attacks on both PASCAL-VOC and NUS-WIDE datasets for one- and two-node attacks on the ML-GCN multi-label classifier. From the results we make the following conclusions: 

\mybullet Using the naive success rate, $SR_N$, all methods have roughly similar performance, yet once easy to detect attacks get filtered out, different methods show very different performance (see $SR_L$ and $SR_G$).  First, notice that the performances have the same trend for the one-node and the two-node attacks. Also, attack success rates on NUS-WIDE are generally lower than PASCAL-VOC as the former dataset has more number of labels and more correlations among them. 

\mybullet The $SR_L$ and $SR_G$ success rate of $\text{GMLA}_\alpha$ is significantly higher than $\text{MLA}_\alpha$ and success rate of $\text{GMLA}_\beta$ is significantly higher than $\text{MLA}_\beta$. This shows the effectiveness of our proposed multi-label attacks that take advantage of label correlations for evading detection. 

\mybullet Notice that the success rates ($SR_L$ and $SR_G$) of $\text{MLA}_\beta$ is significantly low and lower than $\text{MLA}_\alpha$. This is due to the fact that $\text{MLA}_\beta$ fixes all non-target labels and attacks the target labels, which leads to many inconsistency between each target label and its parents and/or children. 

\mybullet Notice that the success rates ($SR_L$ and $SR_G$) of $\text{GMLA}_\beta$ is significantly higher than $\text{GMLA}_\alpha$. This is due to the fact the $\text{GMLA}_\beta$ fixes all labels outside the expanded target set $\Gamma$ while attacking the minimal set $\Gamma$, which preserves consistency not only around the target nodes but in the entire graph. On the other hand, $\text{GMLA}_\alpha$ allows changes of labels outside $\Gamma$, which in some cases lead to inconsistencies in other parts of the graph, leading to detection.

%
%
%

\medskip Figure \ref{fig:attackdetect} shows the local and global detection rates for different attacks on ML-GCN and on the two datasets. From the results, we make the following conclusions: 

\mybullet Generally, detection rates are higher on NUS-WIDE than Pascal-VOC, as there are more inconsistencies caused. This shows that, for a dataset with a large number of labels, it is harder to avoid hierarchical inconsistencies when having a larger number of labels without leveraging effective attacks that take into account label relationships. 

\mybullet Notice that the detection rates of $\text{GMLA}_\beta$ is very close to zero, thanks to ensuring consistency of labels across the entire graph after the attacks. In all cases, each $\text{GMLA}$ type attack has significantly lower detection rate than the corresponding $\text{MLA}$ attack. Also, the detection rate of $\text{GMLA}_\alpha$ increases on NUS-WIDE compared to PASCLA-VOC, as not fixing labels outside $\Gamma$ leads to more inconsistencies due to its larger label set size. 

%

    \begin{table}
    \centering
    \resizebox{1\columnwidth}{!}{%
    \begin{tabular}{|c|c|c|c|c|c|c|c|c|c|c|} 
    \hline 
    \multirow{3}{*}{\textbf{MLR Attacks}} & \multicolumn{10}{c|}{\textbf{ML-GCN}}                                                                                                                             \\ 
    \cline{2-11}
                                          & \multicolumn{5}{c|}{\textbf{Pascal-VOC}}                     & \multicolumn{5}{c|}{\textbf{NUS-WIDE}}                           \\ 
    \cline{2-11}
                                          & $DR_L$        & $DR_G$        & $SR_N$        & $SR_L$        & $SR_G$        & $DR_L$        & $DR_G$        & $SR_N$         & $SR_L$         & $SR_G$          \\ 
    \hline
    MLA$_\alpha$                          & 0.373         & 0.719         & \textbf{1.00} & 0.627         & 0.281         & 0.441         & 0.952         & \textbf{1.00}           & 0.559          & 0.048           \\ 
    \hline
    MLA$_\beta$                           & 0.949         & 0.949         & 0.991         & 0.051         & 0.051         & 0.842         & 0.842         & 0.977          & 0.155          & 0.155           \\ 
    \hline
    GMLA$_\alpha$                         & 0.044         & 0.450         & \textbf{1.00} & 0.956         & 0.550 & 0.193         & 0.843         & 0.998 & 0.805          & 0.157           \\ 
    \hline
    GMLA$_\beta$                          & \textbf{0.00} & \textbf{0.00} & \textbf{1.00} & \textbf{1.00} & \textbf{1.00} & \textbf{0.00} & \textbf{0.00} & 0.996          & \textbf{0.996} & \textbf{0.996}  \\
    \hline 
    \end{tabular}}
    \vspace{-2mm}
    \caption{Performances of MLR single-node attacks ($\epsilon = 0.004$) on ML-GCN model using Pascal-VOC and NUS-WIDE datasets.}
    \label{Tab:tab3}
    \end{table}

    \begin{table}
    \centering
    \resizebox{1\columnwidth}{!}{%
    \begin{tabular}{|c|c|c|c|c|c|c|c|c|c|c|} 
    \hline
    \multirow{3}{*}{\textbf{MLR Attacks}} & \multicolumn{10}{c|}{\textbf{Asymmetric Loss}}                                                                                                         \\ 
    \cline{2-11}
                                          & \multicolumn{5}{c|}{\textbf{PASCAL-VOC}}                                         & \multicolumn{5}{c|}{\textbf{NUS-WIDE}}             \\ 
    \cline{2-11}
                                          & $DR_L$        & $DR_G$        & $SR_N$         & $SR_L$         & $SR_G$         & $DR_L$ & $DR_G$ & $SR_N$         & $SR_L$         & $SR_G$          \\ 
    \hline
    MLA$_\alpha$                          & 0.129         & 0.369         & 0.901          & 0.784          & 0.544          & 0.248  & 0.589  & \textbf{0.996} & 0.749          & 0.409           \\ 
    \hline
    MLA$_\beta$                           & 0.915         & 0.915         & 0.874          & 0.074          & 0.074          & 0.883  & 0.883  & 0.983          & 0.114          & 0.114           \\ 
    \hline
    GMLA$_\alpha$                         & 0.009         & 0.370         & \textbf{0.951} & \textbf{0.942} & 0.599          & 0.047  & 0.369  & 0.990          & 0.943          & 0.625           \\ 
    \hline
    GMLA$_\beta$                          & \textbf{0.00} & \textbf{0.00} & 0.933          & 0.933          & \textbf{0.933} & \textbf{0.00}   & \textbf{0.00}   & 0.982          & \textbf{0.982} & \textbf{0.982}  \\
    \hline 
    \end{tabular}
    }
    \vspace{-2mm}
    \caption{Performances of MLR single-node attacks ($\epsilon = 0.004$) on TResNet-L model trained with Asymmetric loss on two datasets.}
    \label{Tab:tab2}
    \end{table}

    \begin{table}
    \centering
    \resizebox{1\columnwidth}{!}{%
    
    \begin{tabular}{|c|c|c|c|c|c|c|c|c|c|c|} 
    \hline 
    \multirow{3}{*}{\textbf{MLR Attacks}} & \multicolumn{10}{c|}{\textbf{Binary-Relevance}}                                                                                                                      \\ 
    \cline{2-11}
                                          & \multicolumn{5}{c|}{\textbf{Pascal-VOC}}                                         & \multicolumn{5}{c|}{\textbf{NUS-WIDE}}                                            \\ 
    \cline{2-11}
                                          & $DR_L$        & $DR_G$        & $SR_N$         & $SR_L$         & $SR_G$         & $DR_L$        & $DR_G$         & $SR_N$        & $SR_L$         & $SR_G$          \\ 
    \hline
    MLA$_\alpha$        & 0.129         & 0.426         & \textbf{0.998} & 0.869          & 0.572          & 0.274         & 0.795          & \textbf{1.00} & 0.726          & 0.205           \\ 
    \hline
    MLA$_\beta$         & 0.919         & 0.919         & 0.938          & 0.076          & 0.076          & 0.858         & 0.858          & 0.972         & 0.138          & 0.138           \\ 
    \hline
    GMLA$_\alpha$       & 0.012         & 0.439         & 0.971          & \textbf{0.960} & 0.545          & 0.107         & 0.635          & 0.995         & 0.888          & 0.363           \\ 
    \hline
    GMLA$\beta$        & \textbf{0.00} & \textbf{0.00} & 0.941          & 0.941          & \textbf{0.941} & \textbf{0.00} & \textbf{0.000} & 0.991         & \textbf{0.991} & \textbf{0.991}  \\
    \hline 
    \end{tabular}
    }
    \vspace{-2mm}
    \caption{Performances of MLR single-node attacks ($\epsilon = 0.004$) on Binary-Relevance model and two datasets.}
    \label{Tab:tab1}
    \end{table}

\begin{figure}[t]
    \centering
    
    \includegraphics[width=8.4cm]{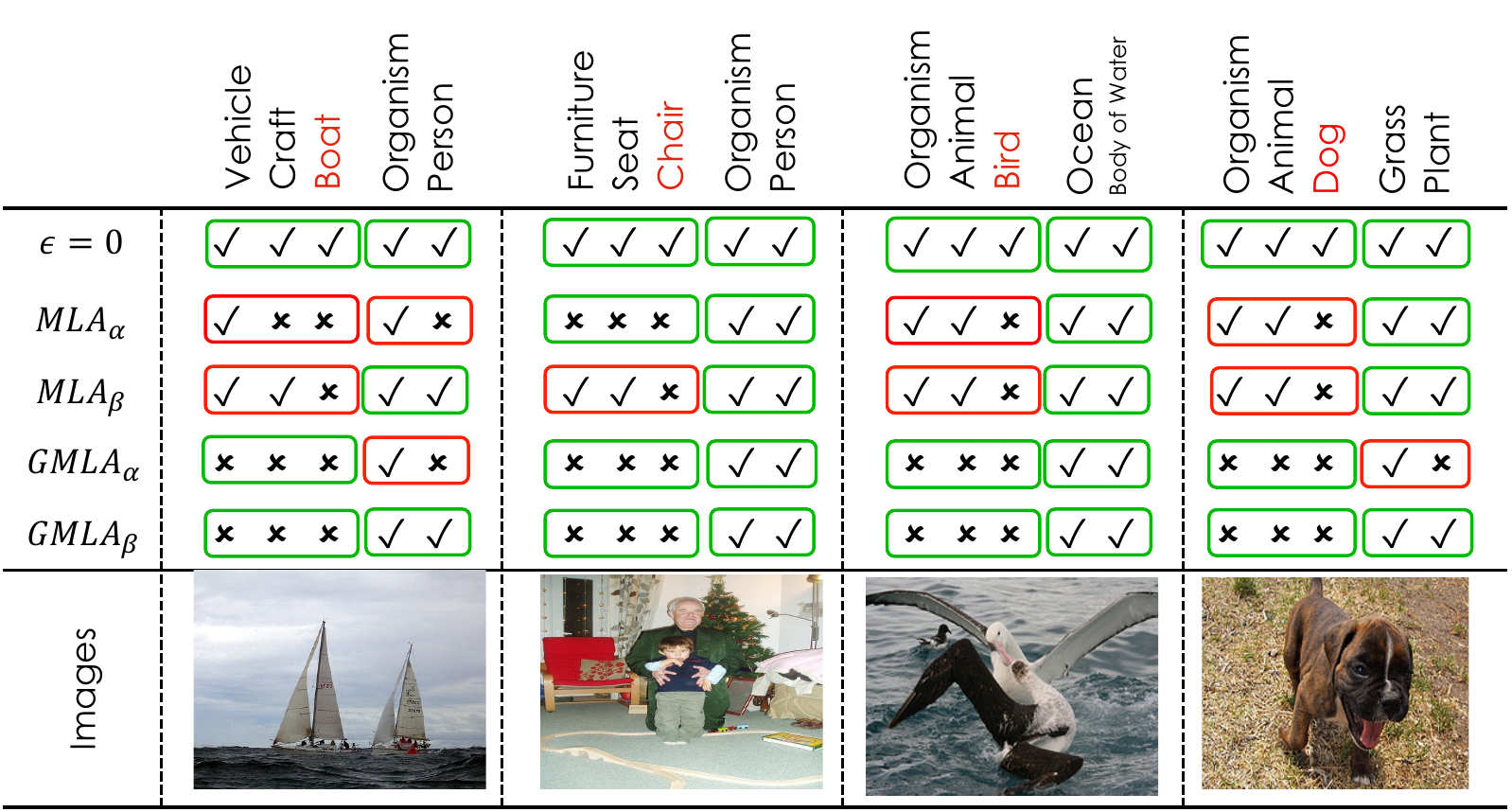}%
    
     \caption{Results of attacking ML-GCN trained on PASCAL-VOC (first two columns) and NUS-WIDE (last two columns). Each column shows the model predictions for clean ($\epsilon = 0$) and attacked images. \emph{Rounded rectangles group hierarchically related labels together}. Inconsistent hierarchies are shown with red 
     rectangles, while consistent hierarchies are shown with green rectangles. The red labels at the top show the ones being attacked.}
     \label{fig:qualresults}
  \end{figure}
  
In Tables \ref{Tab:tab3}, \ref{Tab:tab2}, and \ref{Tab:tab1}, we show the success and detection rates of the four types of attacks for three different multi-label recognition methods. The results are reported for $\epsilon = 0.004$ for which all attacks have achieved high success rate ($SR_N$). Notice that the trends remain similar across different recognition models, where our GMLA attacks achieve higher success rates and much lower detection rates (again, $\text{GMLA}_{\beta}$ has very high success and stays almost undetected independent of the recognition model). Also, notice that MLA attacks are better detected for ML-GCN than the other two models, since it explicitly learns label relationships. 


\mpar{Qualitative Results.} Figure \ref{fig:qualresults} shows qualitative results of attacking ML-GCN using PASCAL-VOC and NUS-WIDE. Notice that in all four and in three cases, respectively, $\text{MLA}_{\beta}$ and $\text{MLA}_{\alpha}$ lead to inconsistencies. For example, in the first image, $\text{MLA}_{\alpha}$ attacks the \emph{boat} label and undesirably affects the \emph{person} label, turning it off while still keeping the label \emph{organism} on, leading to inconsistency. $\text{MLA}_{\beta}$ successfully attacks \emph{boat}, but keeps its parents on, causing inconsistent predictions. For the same first image, $\text{GMLA}_{\alpha}$ keeps consistency locally around \emph{boat}, turning off all its parents as well, but fails to keep consistency for the other part of the graph that contains \emph{person} and \emph{organism}. However, as one can see from the third image, $\text{GMLA}_{\alpha}$, could still keep global consistency (while it is not generally guaranteed). Notice that in all cases, $\text{GMLA}_{\beta}$ successfully modifies the necessary labels to ensure consistency across predictions for all labels.

\section{Conclusions}
We studied the problem of multi-label recognition attacks and proposed a framework for generating image perturbations that respect the relationships among labels according to a knowledge graph. We showed that this leads to our attacks stay undetected but be effective. By extensive experiments on two datasets and using several multi-label recognition models, we show that our method generates extremely successful attacks that, unlike naive multi-label attacks, can evade being detected.

{\small
\bibliographystyle{ieee_fullname}
\bibliography{biblio_bank/adversariallearning,biblio_bank/nlp,biblio_bank/deeplearning,biblio_bank/recognition,biblio_bank/multilabellearning,biblio_bank/object_detection}

\begin{thebibliography}{10}\itemsep=-1pt

\bibitem{Athalye:icml18}
A. Athalye, N. Carlini, and D.~A. Wagner.
\newblock Obfuscated gradients give a false sense of security: Circumventing
  defenses to adversarial examples.
\newblock 2018.

\bibitem{Baruch:arxiv20}
E.~B. Baruch, T. Ridnik, N. Zamir, A. Noy, I. Friedman, M. Protter, and L.
  Zelnik-Manor.
\newblock Asymmetric loss for multi-label classification.
\newblock {\em ArXiv preprint arXiv:2009.14119}, 2020.

\bibitem{Biggio:kdd13}
B. Biggio, I. Corona, D. Maiorca, B. Nelson, N. {\v{S}}rndi{\'{c}}, P. Laskov,
  G. Giacinto, and F. Roli.
\newblock Evasion attacks against machine learning at test time.
\newblock 2013.

\bibitem{Carlini:AISEC17}
N. Carlini and D. Wagner.
\newblock Adversarial examples are not easily detected: Bypassing ten detection
  methods.
\newblock {\em Workshop on Artificial Intelligence and Security}, 2017.

\bibitem{Chen:icme19}
Zhao-Min Chen, Xiu-Shen Wei, Xin Jin, and Yanwen Guo.
\newblock Multi-label image recognition with joint class-aware map
  disentangling and label correlation embedding.
\newblock {\em IEEE International Conference on Multimedia and Expo}, 2019.

\bibitem{Chen:CVPR19}
Z.~M. Chen, X.~S. Wei, P. Wang, and Y. Guo.
\newblock Multi-label image recognition with graph convolutional networks.
\newblock {\em {IEEE} Conference on Computer Vision and Pattern Recognition},
  abs/1904.03582, 2019.

\bibitem{NUS-Wide:09}
T.~S. Chua, J. Tang, R. Hong, H. Li, Z. Luo, and Y.~T. Zheng.
\newblock Nus-wide: A real-world web image database from national university of
  singapore.
\newblock {\em ACM International Conference on Image and Video Retrieval},
  2009.

\bibitem{FCroce:arxiv20}
F. Croce and M. Hein.
\newblock Reliable evaluation of adversarial robustness with an ensemble of
  diverse parameter-free attacks.
\newblock {\em ArXiv}, 2020.

\bibitem{Dao:arxiv21}
S.~D. Dao, E. Zhao, D. Phung, and J. Cai.
\newblock Multi-label image classification with contrastive learning.
\newblock {\em arXiv preprint, arXiv:2107.11626}, 2021.

\bibitem{Dong:cvpr19}
Y. Dong, T. Pang, H. Su, and J. Zhu.
\newblock Evading defenses to transferable adversarial examples by
  translation-invariant attacks.
\newblock {\em {IEEE} Conference on Computer Vision and Pattern Recognition},
  2019.

\bibitem{pascalvoc}
M. Everingham, S.~M.~A. Eslami, L. Van-Gool, C.~K.~I. Williams, J. Winn, and A.
  Zisserman.
\newblock The pascal visual object classes (voc) challenge.
\newblock {\em International Journal of Computer Vision}, 2010.

\bibitem{Eykholt:arxiv18}
K. Eykholt, I. Evtimov, E. Fernandes, B. Li, A. Rahmati, F. Tram{\`e}r, A.
  Prakash, T. Kohno, and D.~X. Song.
\newblock Physical adversarial examples for object detectors.
\newblock {\em arXiv}, 2018.

\bibitem{Eykholt:CVPR18}
K. Eykholt, I. Evtimov, E. Fernandes, B. Li, A. Rahmati, C. Xiao, A. Prakash,
  T. Kohno, and D. Song.
\newblock Robust physical-world attacks on deep learning visual classification.
\newblock {\em {IEEE} Conference on Computer Vision and Pattern Recognition},
  2018.

\bibitem{Feng:AAAI19}
L. Feng, B. An, and S. He.
\newblock Collaboration based multi-label learning.
\newblock {\em {AAAI} Conference on Artificial Intelligence}, 2019.

\bibitem{Goodfellow:ICLR15}
I.~J. Goodfellow, J. Shlens, and C. Szegedy.
\newblock Explaining and harnessing adversarial examples.
\newblock {\em International Conference on Learning Representations}, 2015.

\bibitem{gowal:arxiv19}
Sven Gowal, Jonathan Uesato, Chongli Qin, Po-Sen Huang, Timothy Mann, and
  Pushmeet Kohli.
\newblock An alternative surrogate loss for pgd-based adversarial testing.
\newblock {\em arXiv}, 2019.

\bibitem{Hendrycks:CVPR21}
Dan Hendrycks, Kevin Zhao, Steven Basart, Jacob Steinhardt, and Dawn Song.
\newblock Natural adversarial examples.
\newblock In {\em Proceedings of the IEEE/CVF Conference on Computer Vision and
  Pattern Recognition}, pages 15262--15271, 2021.

\bibitem{Li-Tian:ICCV19}
J.~Li~R. Ji, H. Liu, X. Hong, Y. Gao, and Q. Tian.
\newblock Universal perturbation attack against image retrieval.
\newblock {\em International Conference on Computer Vision}, 2019.

\bibitem{Kingma:ICLR15}
Diederik Kingma and Jimmy Ba.
\newblock Adam: A method for stochastic optimization.
\newblock {\em International Conference on Learning Representations}, 2015.

\bibitem{Kurakin:arXiv16}
A. Kurakin, I. Goodfellow, and S. Bengio.
\newblock Adversarial examples in the physical world.
\newblock {\em ArXiv preprint, arXiv:1607.02533}, 2016.

\bibitem{Kurakin:ICLR17}
A. Kurakin, I. Goodfellow, and S. Bengio.
\newblock Adversarial machine learning at scale.
\newblock {\em International Conference on Learning Representations}, 2017.

\bibitem{OpenImages:16}
A. Kuznetsova, H. Rom, N. Alldrin, J. Uijlings, I. Krasin, J. Pont-Tuset, S.
  Kamali, S. Popov, M. Malloci, A. Kolesnikov, T. Duerig, and V. Ferrari.
\newblock The open images dataset v4: Unified image classification, object
  detection, and visual relationship detection at scale.
\newblock {\em International Journal of Computer Vision}, 2016.

\bibitem{Madry:ICLR18}
A. Madry, A. Makelov, L. Schmidt, D. Tsipras, and A. Vladu.
\newblock Towards deep learning models resistant to adversarial attacks.
\newblock {\em International Conference on Learning Representations}, 2018.

\bibitem{Metzen-Fischer:ICCV19}
J.-H. Metzen, M.-C. Kumar, T. Brox, and V. Fischer.
\newblock Universal adversarial perturbations against semantic image
  segmentation.
\newblock {\em International Conference on Computer Vision}, 2019.

\bibitem{Miller:ACM95}
G.~A. Miller.
\newblock Wordnet: a lexical database for english.
\newblock {\em Communications of the ACM}, 38(11), 1995.

\bibitem{Papernot:essp16}
Nicolas Papernot, Patrick Mcdaniel, Somesh Jha, Matt Fredrikson, Z.~Berkay
  Celik, and Ananthram Swami.
\newblock The limitations of deep learning in adversarial settings.
\newblock {\em IEEE European Symposium on Security and Privacy (EuroS\&P)},
  2016.

\bibitem{Pedersen:naacl04}
T. Pedersen, S. Patwardhan, and J. Michelizzi.
\newblock Wordnet::similarity - measuring the relatedness of concepts.
\newblock 2004.

\bibitem{Ridnik:arxiv20}
T. Ridnik, H. Lawen, A. Noy, and I. Friedman.
\newblock Tresnet: High performance gpu-dedicated architecture.
\newblock {\em ArXiv preprint arXiv:2003.13630}, 2020.

\bibitem{Sharif:CCS16}
M. Sharif, S. Bhagavatula, L. Bauer, and M.~K. Reiter.
\newblock Accessorize to a crime: Real and stealthy attacks on state-of-the-art
  face recognition.
\newblock {\em SIGSAC Conference on Computer and Communications Security},
  2016.

\bibitem{Song:ICDM18}
Q. Song, H. Jin, X. Huang, and X. Hu.
\newblock Multi-label adversarial perturbations.
\newblock {\em IEEE International Conference on Data Mining}, 2018.

\bibitem{Speer:AAAI17}
R. Speer, J. Chin, and C. Havasi.
\newblock Conceptnet 5.5: An open multilingual graph of general knowledge.
\newblock 2017.

\bibitem{Szegedy:ICLR14}
C. Szegedy, W. Zaremba, I. Sutskever, J. Bruna, D. Erhan, I. Goodfellow, and R.
  Fergus.
\newblock Intriguing properties of neural networks.
\newblock {\em International Conference on Learning Representations}, 2014.

\bibitem{Thys:cvprw19}
S. Thys, W.~V. Ranst, and T. Goedemé.
\newblock Fooling automated surveillance cameras: Adversarial patches to attack
  person detection.
\newblock {\em IEEE Conference on Computer Vision and Pattern Recognition
  Workshops}, 2019.

\bibitem{Tsoumakas:IJDWM07}
G. Tsoumakas and I. Katakis.
\newblock Multi-label classification: An overview.
\newblock {\em Intenational Journal Data Warehousing and Mining}, 3, 2007.

\bibitem{wong:arxiv20}
E. Wong, L. Rice, and J.~Z. Kolter.
\newblock Fast is better than free: Revisiting adversarial training.
\newblock {\em arXiv}, 2020.

\bibitem{Jiahao:itm20}
J. Xu, H. Tian, Z. Wang, Y. Wang, W. Kang, and F. Chen.
\newblock Joint input and output space learning for multi-label image
  classification.
\newblock {\em IEEE Transactions on Multimedia}, 2020.

\bibitem{Yang:cvpr16}
H. Yang, J.~T. Zhou, Y. Zhang, B. Gao, J. Wu, and J. Cai.
\newblock Exploit bounding box annotations for multi-label object recognition.
\newblock {\em {IEEE} Conference on Computer Vision and Pattern Recognition},
  2016.

\bibitem{Yang:ecml21}
Z. Yang, Y. Han, and X. Zhang.
\newblock Attack transferability characterization for adversarially robust
  multi-label classification.
\newblock 2021.

\bibitem{Yang:aaai21}
Zhuo Yang, Yufei Han, and Xiangliang Zhang.
\newblock Characterizing the evasion attackability of multi-label classifiers.
\newblock 2021.

\bibitem{Ye:ECCV20}
J. Ye, J. He, X. Peng, W. Wu, and Y. Qiao.
\newblock Attention-driven dynamic graph convolutional network for multi-label
  image recognition.
\newblock {\em European Conference on Computer Vision}, 2020.

\bibitem{yu:cvpr21}
Y. Yu, X. Gao, and C. Xu.
\newblock Lafeat: Piercing through adversarial defenses with latent features.
\newblock {\em {IEEE} Conference on Computer Vision and Pattern Recognition},
  2021.

\bibitem{Zheng:arxiv18}
Tianhang Zheng, Changyou Chen, and Kui Ren.
\newblock Distributionally adversarial attack.
\newblock {\em AAAI Conference on Artificial Intelligence}, 2018.

\bibitem{Zhou:ijcnn21}
N. Zhou, W. Luo, J. Zhang, L. Kong, and H. Zhang.
\newblock Hiding all labels for multi-label images: An empirical study of
  adversarial examples.
\newblock {\em International Joint Conference on Neural Networks}, 2021.

\bibitem{Zgner:sigkdd18}
D. Z{\"u}gner, A. Akbarnejad, and S. G{\"u}nnemann.
\newblock Adversarial attacks on neural networks for graph data.
\newblock {\em International Conference on Knowledge Discovery \& Data Mining},
  2018.

\end{thebibliography}
}

\end{document}